\documentclass[10pt,twocolumn,letterpaper]{article}

\usepackage{cvpr}
\usepackage{times}
\usepackage{epsfig}
\usepackage{graphicx}
\usepackage{amsmath}
\usepackage{amssymb}

\usepackage{caption}
\usepackage{enumerate}
\usepackage[ruled]{algorithm2e}
\usepackage[table,xcdraw]{xcolor}
\usepackage{color}
\usepackage{enumitem}
\usepackage[numbers]{natbib}
\usepackage{float}
\usepackage{subfigure}
\usepackage{threeparttable}
\usepackage{multirow}
\usepackage{booktabs}
\usepackage[table]{xcolor}
\usepackage{arydshln}
\usepackage{threeparttable}
\usepackage{colortbl}
\usepackage{enumitem}
\usepackage[numbers]{natbib}
\usepackage{bm}
\usepackage{array}
\usepackage[table]{xcolor}
\usepackage{colortbl}
\usepackage{siunitx}
\usepackage{url}

\usepackage[pagebackref=true,breaklinks=true,letterpaper=true,colorlinks,bookmarks=false]{hyperref}
\cvprfinalcopy 

\def\cvprPaperID{2917} 

\ifcvprfinal\fi
\begin{document}

\title{LiDAR-based Online 3D Video Object Detection with Graph-based Message Passing and Spatiotemporal Transformer Attention}

\author{Junbo Yin\textsuperscript{1,2}, Jianbing Shen\textsuperscript{1,4}\thanks{Corresponding author: \textit{Jianbing Shen}.}, 
Chenye Guan\textsuperscript{2,3}, Dingfu Zhou\textsuperscript{2,3}, Ruigang Yang\textsuperscript{2,3,5}\\
\small{\quad$^1$}\small Beijing Lab of Intelligent Information Technology,
School of Computer Science, Beijing Institute of Technology, China \\ 
\small{$^2$} \small Baidu Research \quad \small{$^3$} \small National Engineering Laboratory of Deep Learning Technology and Application, China \quad \\
\small{$^4$} \small Inception Institute of Artificial Intelligence, UAE \quad
\small{$^5$} \small University of Kentucky, Kentucky, USA\\
{\tt\small \{yinjunbocn, shenjianbingcg\}@gmail.com} \quad
{\tt\small \url{https://github.com/yinjunbo/3DVID}}
}

\maketitle
\thispagestyle{empty}

\begin{abstract}
Existing LiDAR-based 3D object detectors usually focus on the single-frame detection, while ignoring the spatiotemporal information in consecutive point cloud frames. In this paper, we propose an end-to-end online 3D video object detector that operates on point cloud sequences. The proposed model comprises a spatial feature encoding component and a spatiotemporal feature aggregation component. 
In the former component, a novel Pillar Message Passing Network (PMPNet) is proposed to encode each discrete point cloud frame. It adaptively collects information for a pillar node from its neighbors by iterative message passing, which effectively enlarges the receptive field of the pillar feature. In the latter component, we propose an Attentive Spatiotemporal Transformer GRU (AST-GRU) to aggregate the spatiotemporal information, which enhances the conventional ConvGRU with an attentive memory gating mechanism. AST-GRU contains a Spatial Transformer Attention (STA) module and a Temporal Transformer Attention (TTA) module, which can emphasize the foreground objects and align the dynamic objects, respectively. Experimental results demonstrate that the proposed 3D video object detector achieves state-of-the-art performance on the large-scale nuScenes benchmark.
\end{abstract}



\begin{figure}
\begin{center}
\includegraphics[width=0.475\textwidth]{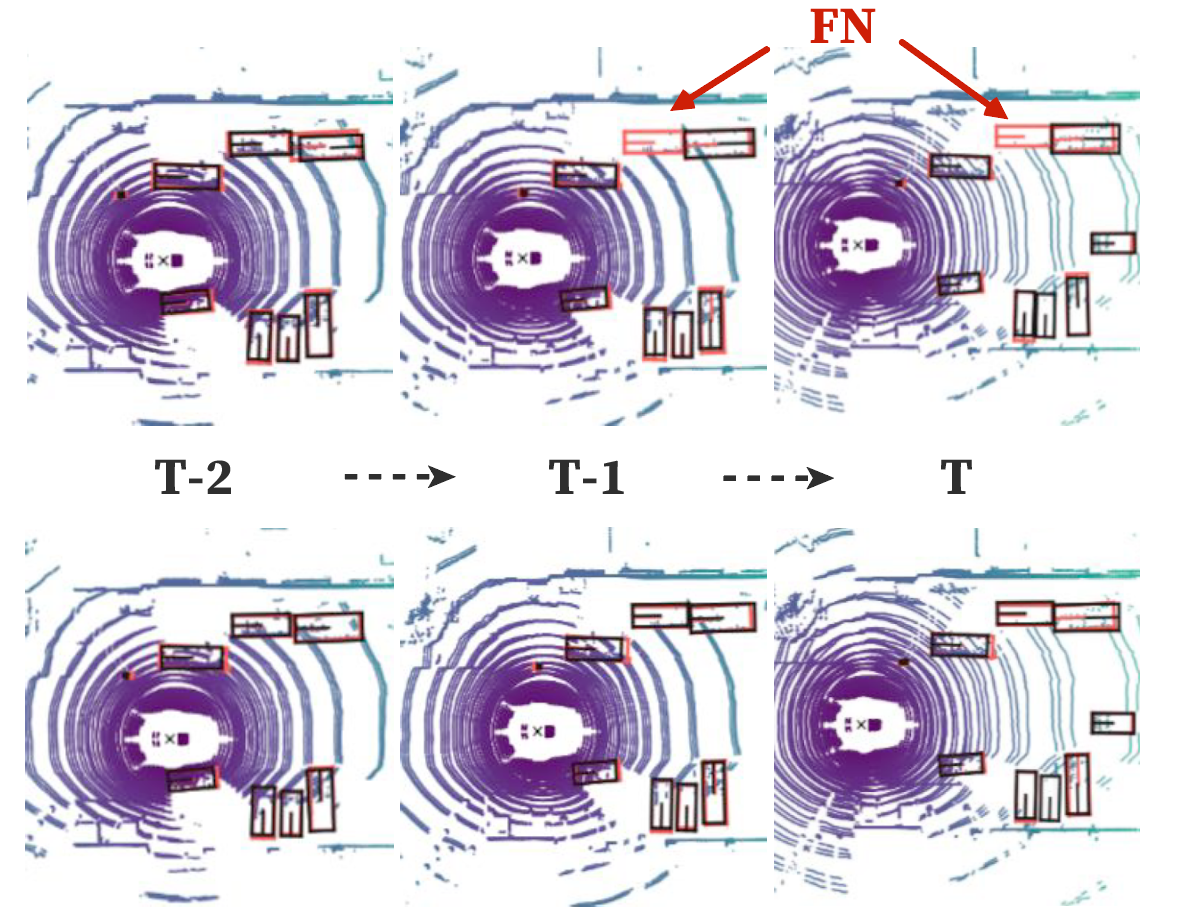}
\end{center}
\vspace{-18pt}
\caption{\small {\textbf{Occlusion situation in autonomous driving scenarios.} Typical single-frame 3D object detector, \eg \cite{lang2019pointpillars}, often leads to false-negative (FN) results (top row). In contrast, our online 3D video object detector can handle this (bottom row). The grey and red boxes denote the predictions and ground-truths, respectively.}}
\label{fig:video}
\vspace{-4mm}
\end{figure}
\section{Introduction}
LiDAR-based 3D object detection plays a critical role in a wide range of applications, such as autonomous driving, robot navigation and virtual/augmented reality~\cite{geiger2012we,song2019apollocar3d}. The majority of current 3D object detection approaches~\cite{shi2019pointrcnn, yang2019std, chen2019fast, zhou2018voxelnet, lang2019pointpillars} follow the \textit{single-frame} detection paradigm, while few of them perform detection in the point cloud \textit{video}. A point cloud video is defined as a temporal sequence of point cloud frames. For instance, in the nuScenes dataset~\cite{caesar2019nuScenes}, $20$ point cloud frames can be captured per second with a modern 32-beam LiDAR sensor. Detection in {single frame} may suffer from several limitations due to the sparse nature of point cloud. In particular, occlusions, long-distance and non-uniform sampling inevitably occur on a certain frame, where a single-frame object detector is incapable of handling these situations, leading to a deteriorated performance, as shown in Fig~\ref{fig:video}. However, a point cloud video contains rich spatiotemporal information of the foreground objects, which can be explored to improve the detection performance. 
The major concern of constructing a 3D video object detector is how to model the spatial and temporal feature representation for the consecutive point cloud frames. In this work, we propose to integrate a graph-based spatial feature encoding component with an attention-aware spatiotemporal feature aggregation component, to capture the video coherence in consecutive point cloud frames, which yields an end-to-end online solution for the LiDAR-based 3D video object detection.

 Popular single-frame 3D object detectors tend to first discretize the point cloud into voxel or pillar girds~\cite{zhou2018voxelnet,yan2018second,lang2019pointpillars}, and then extract the point cloud features using stacks of convolutional neural networks (CNNs). Such approaches incorporate the success of existing 2D or 3D CNNs and usually gain better computational efficiency compared with the point-based methods~\cite{shi2019pointrcnn,qi2019deep}. Therefore, in our spatial feature encoding component, we also follow this paradigm to extract features for each input frame. However, a potential problem with these approaches lies in that they only focus on a locally aggregated feature, \ie, employing a PointNet~\cite{qi2017pointnet} to extract features for \textit{separate} voxels or pillars as in~\cite{zhou2018voxelnet} and~\cite{lang2019pointpillars}. To further enlarge the receptive fields, they have to apply the stride or pooling operations repeatedly, which will cause the loss of the spatial information. To alleviate this issue, we propose a novel graph-based network, named Pillar Message Passing Network (PMPNet), which treats a \textit{non-empty} pillar as a graph node and adaptively enlarges the receptive field for a node by aggregating messages from its neighbors. PMPNet can mine the rich geometric relations among different pillar grids in a discretized point cloud frame by iteratively reasoning on a $k$-NN graph. This effectively encourages information exchanges among different spatial regions within a frame.

After obtaining the spatial features of each input frame, we assemble these features in our spatiotemporal feature aggregation component. Since ConvGRU~\cite{ballas2016delving} has shown promising performance in the 2D video understanding field, we suggest an Attentive Spatiotemporal Transformer GRU (AST-GRU) to extend ConvGRU to the 3D field through capturing dependencies of consecutive point cloud frames with an attentive memory gating mechanism. Specifically, there exist two potential limitations when considering the LiDAR-based 3D video object detection in autonomous driving scenarios. First, in the bird's eye view, most foreground objects (\eg, cars and pedestrians) occupy small regions, and the background noise is inevitably accumulated as computing the new memory in a recurrent unit. Thus, we propose to exploit the Spatial Transformer Attention (STA) module, an intra-attention derived from \cite{vaswani2017attention, wang2018non}, to suppress the background noise and emphasize the foreground objects by attending each pixel with the context information. Second, when updating the memory in the recurrent unit, the spatial features of the two inputs (\ie, the old memory and the new input) are not well aligned. In particular, though we can accurately align the static objects across frames using the ego-pose information, the dynamic objects with large motion are not aligned, which will impair the quality of the new memory. To address this, we propose a Temporal Transformer Attention (TTA) module that adaptively captures the object motions in consecutive frames with a temporal inter-attention mechanism. This will better utilize the modified deformable convolutional layers~\cite{zhu2019deformable,zhu2019empirical}. Our AST-GRU can better handle the spatiotemporal features and produce a more reliable new memory, compared with the vanilla ConvGRU. To summarize, we propose a new LiDAR-based online 3D video object detector that leverages the previous long-term information to improve the detection performance. In our model, a novel PMPNet is introduced to adaptively enlarge the receptive field of the pillar nodes in a discretized point clod frame by iterative graph-based message passing. The output sequential features are then aggregated in the proposed AST-GRU to mine the rich coherence in the point cloud video by using an attentive memory gating mechanism. Extensive evaluations demonstrate that our 3D \textit{video} object detector achieves better performance against the \textit{single-frame} detectors on the large-scale nuScenes benchmark.


\begin{figure*}[t]
  \centering
      \includegraphics[width=0.95\linewidth]{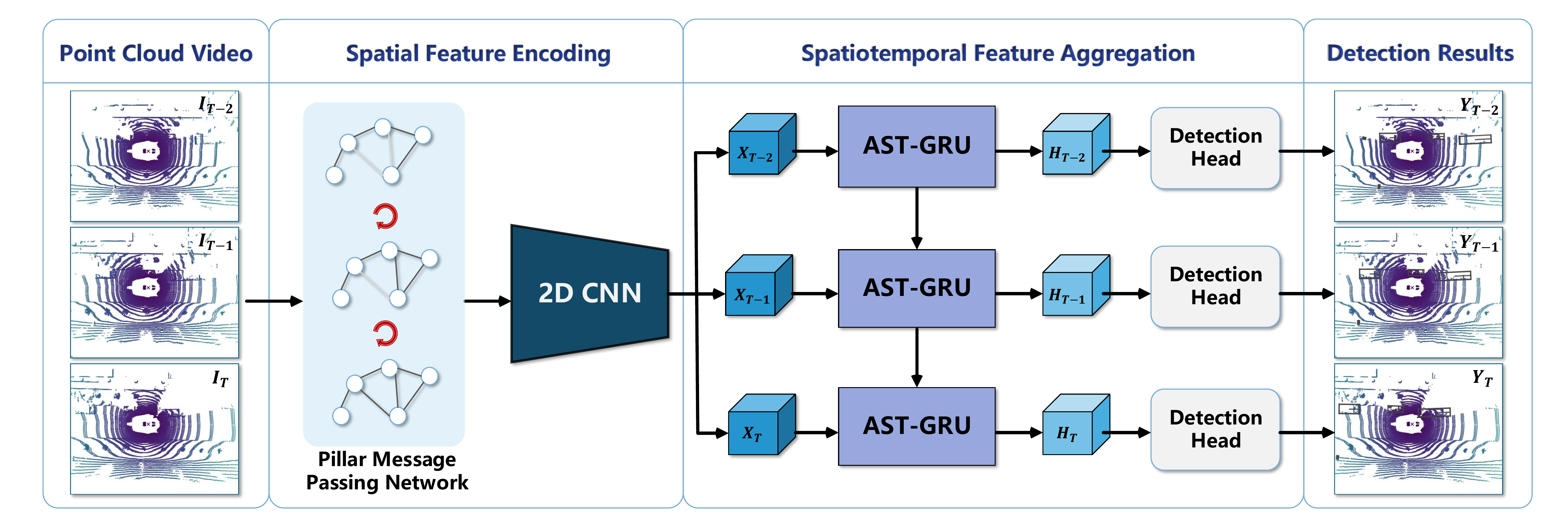}
\vspace{-4pt}
\caption{\small \textbf{Our online 3D video object detection framework} includes a spatial feature encoding component and a spatiotemporal feature aggregation component. 
In the former component, a novel PMPNet (\S\ref{subsec:PMPN}) is proposed to extract the spatial features of each point cloud frame. Then, features from consecutive frames are sent to the AST-GRU (\S\ref{subsec:ASTGRU}) in the latter component, to aggregate the spatiotemporal information with an attentive memory gating mechanism.}
\label{fig:framework}
\vspace{-4mm}
\end{figure*}
\section{Related Work}
\noindent\textbf{LiDAR-based 3D Object Detection.} Existing works on 3D object detection can be roughly categorized into three groups, which are LiDAR-based~\cite{shi2019pointrcnn, yang2019std, zhou2018voxelnet, lang2019pointpillars, zhou2019iou, yan2018second}, image-based~\cite{ku2019monocular, wang2019pseudo, li2019stereo, manhardt2019roi, li2019gs3d} and multi-sensor fusion-based~\cite{chen2017multi, liang2019multi, liang2018deep, ku2018joint, qi2018frustum} methods. Here, we focus on the LiDAR-based approaches since they are less sensitive to different illumination and weather conditions. Among them, one category~\cite{zhou2018voxelnet,yang2018pixor,lang2019pointpillars} typically discretizes the point cloud into regular girds (\eg, voxels or pillars), and then exploits the 2D or 3D CNNs for features extraction. Another category~\cite{shi2019pointrcnn, yang2019std, chen2019fast} learns 3D representations directly from the original point cloud with a point-wise feature extractor like PointNet++~\cite{qi2017pointnet}. It is usually impractical to directly apply the point-based detectors in scenes with large-scale point clouds, for they tend to perform feature extraction for every single point. For instance, a keyframe in nuScenes dataset~\cite{caesar2019nuScenes} contains {300,000} point clouds, which are densified by 10 non-keyframe LiDAR sweeps within \SI{0.5}{s}. Operating on point clouds with such a scale will lead to non-trivial computation cost and memory demand. In contrast, the voxel-based methods can tackle this kind of difficulty for they are less sensitive to the number of points. Zhou~\textit{et al.}~\cite{zhou2018voxelnet} first apply the end-to-end CNNs for voxel-based 3D object detection. They propose to describe each voxel with a Voxel Feature Encoding (VFE) layer, and utilize cascade 3D and 2D CNNs to extract the deep features. Then a Region Proposal Network (RPN) is employed to obtain the final detection results. 
After that, Lang \textit{et al.}~\cite{lang2019pointpillars} further extend~\cite{zhou2018voxelnet} by projecting the point clouds to the the bird's eye view and encoding each discretized gird (named pillars) with a Pillar Feature Network (PFN).

Both the VFE layers and the PFN only take into account \textit{separate} voxels or pillars when generating the grid-level representation, which ignores the information exchange in larger spatial regions. In contrast, our PMPNet encodes the pillar feature from a global perspective by graph-based message passing, and thus promotes the representation with the non-local property. 
Besides, all these \textit{single-frame} 3D object detectors can only process the point cloud data frame-by-frame, lacking the exploration of the temporal information. Though~\cite{luo2018fast} applies temporal 3D ConvNet on point cloud sequences, it encounters the feature collapse issue when downsampling the features in the temporal domain. Moreover, it cannot deal with long-term sequences with multi-frame labels. Our AST-GRU instead captures the long-term temporal information with an attentive memory gating mechanism, which can fully mine the spatiotemporal coherence in the point cloud video. 

\noindent\textbf{Graph Neural Networks.} 
Graph Neural Networks (GNNs) are first introduced by Gori~\textit{et al.}~\cite{gori2005new} to model the intrinsic relationships of the graph-structured data. Then Scarselli~\textit{et al.}~\cite{scarselli2008graph} extend it to different types of graphs.
Afterward, GNNs are explored in two directions in terms of different message propagation strategies. The first group~\cite{li2016gated, kearnes2016molecular, zayats2018conversation, peng2017cross, qi2018learning} uses the gating mechanism to enable the information to propagate across the graph. For instance, Li~\textit{et al.}~\cite{li2016gated} leverage the recurrent neural networks to describe the state of each graph node. Then, Gilmer~\textit{et al.}~\cite{gilmer2017neural} generalizes a framework to formulate the graph reasoning as a parameterized message passing network. Another group~\cite{bruna2014spectral,henaff2015deep,defferrard2016convolutional,hu2020infinitely,li2020self} integrates convolutional networks to the graph domain, named as Graph Convolutional Neural Networks (GCNNs), which update node features via stacks of graph convolutional layers. GNNs have achieved promising results in many areas~\cite{defferrard2016convolutional, fan2019understanding, wang2018attentive, battaglia2016interaction, wang2020hierarchical} due to the great expressive power of graphs. Our PMPNet belongs to the first group by capturing the pillar features with a gated message passing strategy, which is used to construct the spatial representation for each point cloud frame.


\section{Model Architecture}
\vspace{-2pt}
In this section, we elaborate on our online 3D video object detection framework. As shown in Fig.~\ref{fig:framework}, it consists of a spatial feature encoding component and a spatiotemporal feature aggregation component. Given the input sequences $\{\bm{I}_t\}_{t=1}^{T}$ with $T$ frames, we first convert the point cloud coordinates from the previous frames $\{\bm{I}_t\}_{t=1}^{T-1}$ to the current frame $\bm{I}_T$ using the GPS data, so as to eliminate the influence of the ego-motion and align the static objects across frames. Then, in the spatial feature encoding component, we extract features for each frame with the Pillar Message Passing Network (PMPNet) (\S\ref{subsec:PMPN}) and a 2D backbone, producing sequential features $\{\bm{X}_t\}_{t=1}^{T}$. After that, these features are fed into the Attentive Spatiotemporal Transformer Gated Recurrent Unit (AST-GRU) (\S\ref{subsec:ASTGRU}) in the spatiotemporal feature aggregation component, to generate the new memory features $\{\bm{H}_t\}_{t=1}^{T}$.
Finally, a RPN head is applied on $\{\bm{H}_t\}_{t=1}^{T}$ to give the final detection results $\{\bm{Y}_t\}_{t=1}^{T}$. Some network architecture details are provided in \S\ref{subsec:detail}.

\begin{figure}
\begin{center}
\includegraphics[width=0.48\textwidth]{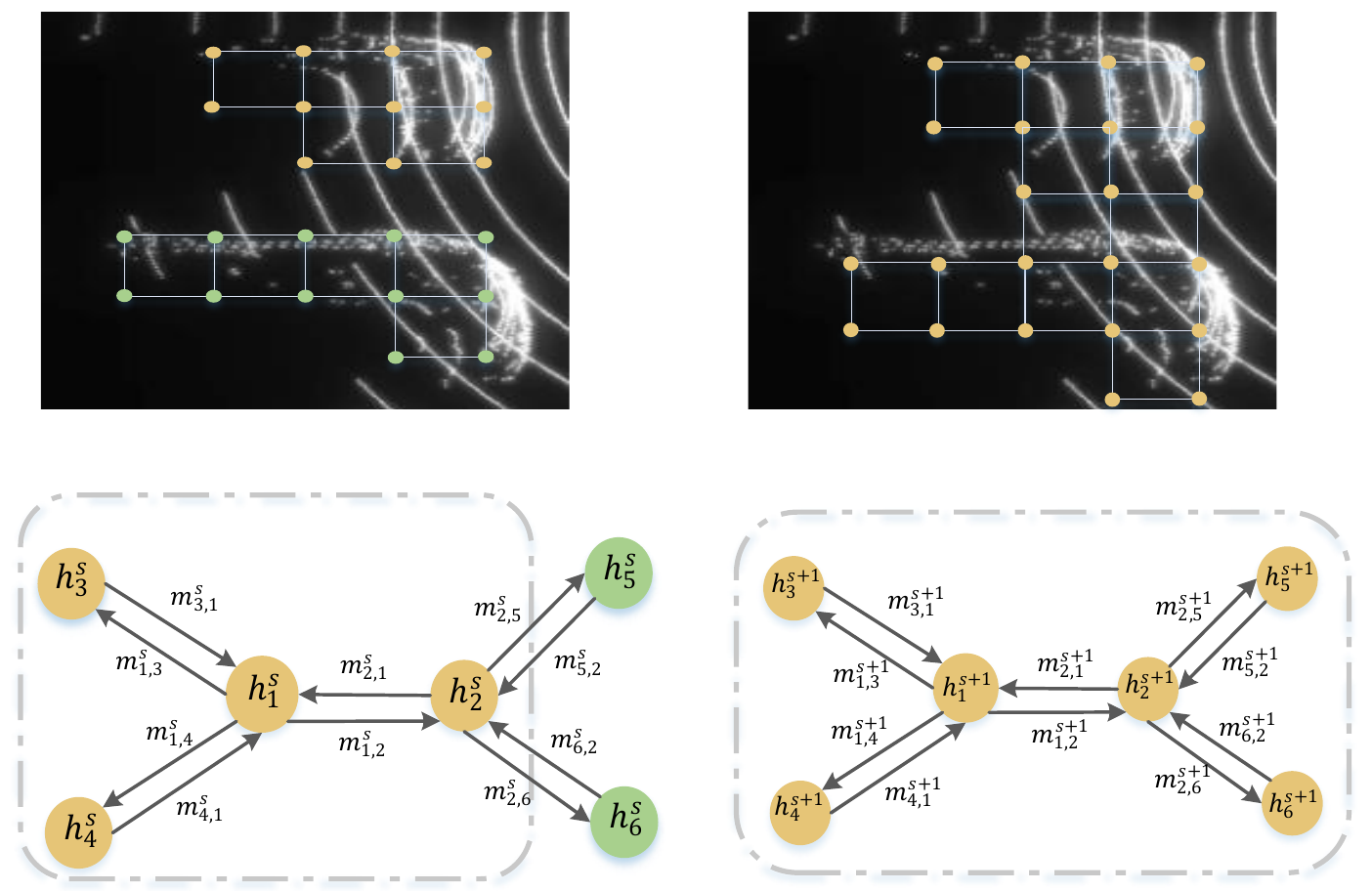}
\end{center}
\vspace{-18pt}
\caption{\small {\textbf{Illustration of one iteration step for message propagation}, where $h_i$ is the state of node $v_i$. In step $s$, the neighbors for $h_1$ are $\{h2, h3, h4\}$ (within the gray dash line), presenting the pillars in the top car. After aggregating messages from the neighbors, the receptive field of $h_1$ is enlarged in step $s+1$, indicating the relations with nodes from the bottom car are modeled.}}
\label{fig:gnn}
\vspace{-4mm}
\end{figure}
\subsection{Pillar Message Passing Network}
\label{subsec:PMPN} 
Previous point cloud encoding layers (\eg, the VFE layers in~\cite{zhou2018voxelnet} and the PFN in~\cite{lang2019pointpillars}) for voxel-based 3D object detection typically encode each voxel or pillar separately, which limits the expressive power of the grid-level representation due to the small receptive field of each local grid region. Our PMPNet instead seeks to explore the rich spatial relations among different gird regions by treating the {non-empty} pillar grids as graph nodes. Such design effectively reserves the non-Euclidean geometric characteristics of the original point clouds and enhance the output pillar features with a non-locality property.

Given an input point cloud frame $\bm{I}_t$, we first uniformly discretize it into a set of pillars $\mathcal{P}$, with each pillar uniquely associated with a spatial coordinate in the x-y plane as in~\cite{lang2019pointpillars}. Then, PMPNet maps the resultant pillars to a directed graph $\mathcal{G}=(\mathcal{V},\mathcal{E})$, where node $v_i \in \mathcal{V}$ represents a {non-empty} pillar $P_i\in\mathcal{P}$ and edge ${e}_{i, j}\in\mathcal{E}$ indicates the message passed from node $v_i$ to $v_j$. For reducing the computational overhead, we define $\mathcal{G}$ as a $k$-nearest neighbor ($k$-NN) graph, which is built from the geometric space by comparing the centroid distance among different pillars. 

To explicitly mine the rich relations among different pillar nodes, PMPNet performs iterative message passing on $\mathcal{G}$ and updates the nodes state at each iteration step. 
Concretely, given a node $v_i$, we first utilize a pillar feature network (PFN)~\cite{lang2019pointpillars} to describe its initial state $\bm{h}_i^0$ at iteration step $s=0$:
\begin{equation}
\begin{aligned}
\label{eq:pfn}
\bm{h}_i^0 = F_{\text{PFN}}{(P_i)}\in\mathbb{R}^{L},
\end{aligned}
\end{equation}
where $\bm{h}_i^0$ is a $L$-dim vector and $P_i \in \mathbb{R}^{N \times D}$ presents a pillar containing $N$ LiDAR points, with each point parameterized by $D$ dimension representation (\eg, the XYZ coordinates and the received reflectance). The PFN is realized by applying fully connected layers on each point within the pillar, then summarizing features of all points through a channel-wise maximum operation. The initial node state $\bm{h}_i^0$ is a locally aggregated feature, only including points information within a certain pillar grid. 
      
Next, we elaborate on the message passing process.
One iteration step of message propagation is illustrated in Fig.~\ref{fig:gnn}. At step $s$, a node $v_i$ aggregates information from all the neighbor nodes $v_j \in \bm{\Omega}_{v_i}$ in the $k$-NN graph. We define the incoming edge feature from node $v_j$ as $\bm{e}_{j, i}^{s}$, indicating the relation between node $v_i$ and $v_j$. Inspired by \cite{wang2019dynamic}, the incoming edge feature $\bm{e}_{j, i}^{s}$ is given by:
\begin{equation}
\begin{aligned}
\label{eq:edge}
\bm{e}_{j, i}^{s} = \bm{h}_j^s - \bm{h}_i^s\in\mathbb{R}^{L},
\end{aligned}
\end{equation}       
 which is an asymmetric function encoding the local neighbor information. Accordingly, we have the message passed from $v_j$ to $v_i$, which is denoted as:
\begin{equation}
\begin{aligned}
\label{eq:message1}
\bm{m}_{j, i}^{s+1} = \phi_{\theta}{([\bm{h}_i^s, \bm{e}_{j, i}^s])}\in\mathbb{R}^{L'},
\end{aligned}
\end{equation}      
where $\phi_{\theta}$ is parameterized by a fully connected layer, which takes as input the concatenation of $\bm{h}_i^s$ and $\bm{e}_{j, i}^s$, and yields a $L'$-dim feature. 

After computing all the pair-wise relations between $v_i$ and the neighbors ${v_j}\in{\bm{\Omega}_{v_i}}$ of , we summarize the received $k$ messages with a maximum operation:
\begin{equation}
\begin{aligned}
\label{eq:message2}
\bm{m}_{i}^{s+1} = \mathop {\text{max}}\limits_{j\in\bm{\Omega}_i}{(\bm{m}_{j, i}^{s+1})}\in\mathbb{R}^{L'},
\end{aligned}
\end{equation}
Then, we update the node state $\bm{h}_i^s$ with $\bm{h}_i^{s+1}$ for node $v_i$.
 The update process should consider both the newly collected message $\bm{m}_{i}^{s+1}$ and the previous state $\bm{h}_i^s$. Recurrent neural network and its variants~\cite{hochreiter1997long,sutskever2014sequence} can adaptively capture dependencies in different time steps. Hence, we utilize Gated Recurrent Unit (GRU)~\cite{cho2014learning} as the update function for its better convergence characteristic. The update process is then formulated as follows:
\begin{equation}
\begin{aligned}
\label{eq:update}
\bm{h}_{i}^{s+1} = {\text{GRU}}(\bm{h}_i^s, \bm{m}_{i}^{s+1})\in\mathbb{R}^{L},
\end{aligned}
\end{equation}
In this way, the new node state $\bm{h}_i^{s+1}$ contains the information from all the neighbor nodes of $v_i$. Moreover, a neighbor node $v_j$ also collects information from its own neighbors $\bm{\Omega}_{v_j}$. Consequently, after the totally $S$ iteration steps, node $v_i$ is able to aggregate information from the high-order neighbors. This effectively enlarges the perceptual range for each pillar grid and enables our model to better recognize objects from a global view.
  
Note that each pillar corresponds with a spatial coordinate in the x-y plane. Therefore, after performing the iterative message passing, the encoded pillar features are then scattered back as a 3D tensor $\bm{\tilde{I}}_t\in\mathbb{R}^{W\times H\times C}$, which can be further exploited by the 2D CNNs. 
Here, we leverage the backbone network in~\cite{zhou2018voxelnet} to further extract features for $\bm{\tilde{I}}_t$:
\begin{equation}
\begin{aligned}
\label{eq:2dbackbone}
\bm{X}_{t} = {F_{\text{B}}}(\bm{\tilde{I}}_t)\in\mathbb{R}^{w\times h\times c},
\end{aligned}
\end{equation}
where $F_{\text{B}}$ denotes the backbone network and $\bm{X}_{t}$ is the spatial features of $\bm{{I}}_t$. Details of the PMPNet and the backbone network can be found in \S\ref{subsec:detail}. 
\vspace{-1pt}
\subsection{Attentive Spatiotemporal Transformer GRU}
\begin{figure}
\begin{center}
\includegraphics[width=0.48\textwidth]{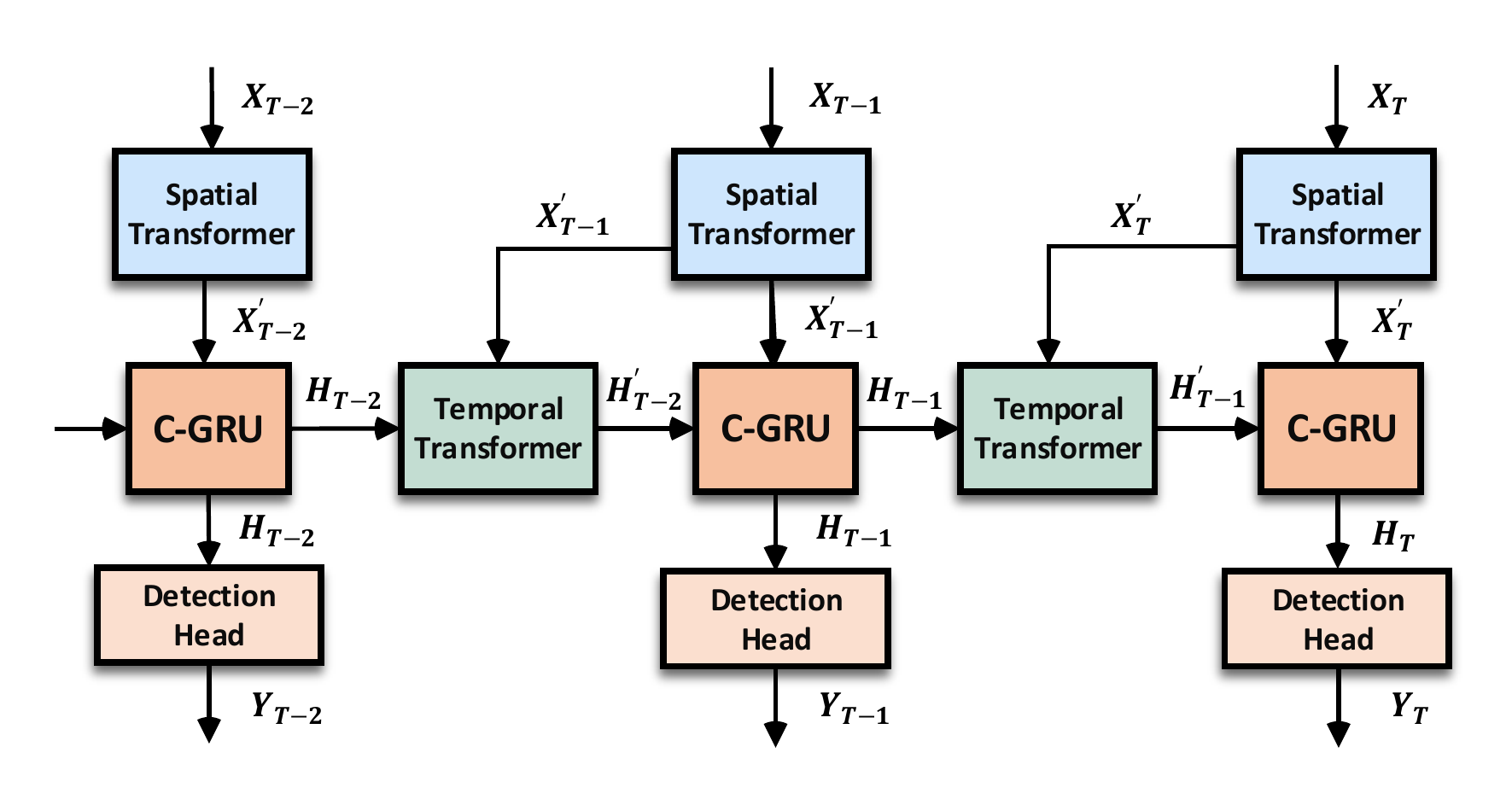}
\end{center}
\vspace{-18pt}
\caption{\small {\textbf{The detailed architecture of the proposed AST-GRU}, which consists of a spatial transformer attention (STA) module and a temporal transformer attention (TTA) module. AST-GRU models the dependencies of consecutive frames and produces the attentive new memory $\{\bm{H}_t\}_{t=1}^{T}$.}}
\label{fig:GRU}
\vspace{-4mm}
\end{figure}

\vspace{-1pt}
\label{subsec:ASTGRU}
Since the sequential features $\{\bm{X}_t\}_{t=1}^{T}$ produced by the spatial feature encoding component are regular tensors, we can employ the ConvGRU~\cite{ballas2016delving} to fuse these features in our spatiotemporal feature aggregation component. However, it may suffer from two limitations when directly applying the ConvGRU. On the one hand, the interest objects are relatively small in the bird's eye view compared with those in the 2D images (\eg, an average of $18\times8$ pixels for cars with the pillar size of $0.25^2~m^2$). This may cause the background noise to dominate the results when computing the memory. On the other hand, though the static objects can be well aligned across frames using the GPS data, the dynamic objects with large motion still lead to an inaccurate new memory. To address the above issues, we propose the AST-GRU to equip the vanilla ConvGRU~\cite{ballas2016delving} with a spatial transformer attention (STA) module and a temporal transformer attention (TTA) module. 
As illustrated in Fig.~\ref{fig:GRU}, the STA module stresses the foreground objects in $\{\bm{X}_t\}_{t=1}^{T}$ and produces the attentive new input $\{\bm{X}_t^{'}\}_{t=1}^{T}$, while the TTA module aligns the dynamic objects in  $\{\bm{H}_{t-1}\}_{t=1}^{T}$ and $\{\bm{X}_t^{'}\}_{t=1}^{T}$, and outputs the attentive old memory $\{\bm{H}_{t-1}^{'}\}_{t=1}^{T}$. Then, $\{\bm{X}_t^{'}\}_{t=1}^{T}$ and $\{\bm{H}_{t-1}^{'}\}_{t=1}^{T}$ are used to generate the new memory $\{\bm{H}_t\}_{t=1}^{T}$, and further produce the final detections $\{\bm{Y}_t\}_{t=1}^{T}$. Before giving the details of the STA and TTA modules, we first review the vanilla ConvGRU.

\noindent\textbf{Vanilla ConvGRU.}  
GRU model~\cite{cho2014learning} operates on a sequence of inputs to adaptively capture the dependencies in different time steps with a memory mechanism. ConvGRU is a variant of the conventional GRU model, which employs convolution operations rather than the fully connected ones, to reduce the number of parameters and preserve the spatial resolution of the input features. ConvGRU has made promising results on many tasks~\cite{liu2018mobile, wang2019zero, lai2019video, wang2019iterative}, and has shown better results than the LSTM~\cite{xingjian2015convolutional} counterparts in terms of the convergence time~\cite{chung2014empirical}. More specifically, ConvGRU contains an update gate $\bm{z}_t$, a reset gate $\bm{r}_t$, a candidate memory $\tilde{\bm{H}}_t$ and a new memory $\bm{H}_t$. At each time step, the new memory $\bm{H}_t$ (also named as the hidden state) is computed based on the old memory $\bm{H}_{t-1}$ and the new input $\bm{X}_t$, which can be denoted by the following equations: 
\begin{flalign}
\label{eq:gru}
&\bm{z}_{t} = \sigma(\bm{W}_{z}*\bm{X}_{t}+\bm{U}_{z}*\bm{H}_{t-1}),\\
&\bm{r}_{t} = \sigma(\bm{W}_{r}*\bm{X}_{t}+\bm{U}_{r}*\bm{H}_{t-1}),\\
&\tilde{\bm{H}}_{t} = \tanh(\bm{W}*\bm{X}_{t}+\bm{U}*(\bm{r}_{t}\circ\bm{H}_{t-1})),\\
&\bm{H}_{t} = (\bm{1}-\bm{z}_t)\circ\bm{H}_{t-1}+\bm{z}_t\circ\tilde{\bm{H}}_{t},
\end{flalign}
where `*' and `$\circ$' denote the convolution operation and Hadamard product, and $\sigma$ is a sigmoid function. $\bm{W}, \bm{W}_{z},\bm{W}_{r}$ and $\bm{U}, \bm{U}_{z},\bm{U}_{r}$ are the 2D convolutional kernels. When computing the candidate memory $\tilde{\bm{H}}_{t}$, the importance of the old memory $\bm{H}_{t-1}$ and the new input $\bm{X}_t$ is determined by the reset gate $\bm{r}_t$, \ie, the information of $\tilde{\bm{H}}_{t}$ all comes from $\bm{X}_t$ when $\bm{r}_{t}=0$. Additionally, the update gate $\bm{z}_t$ decides the degree to which the unit accumulates the old memory $\bm{H}_{t-1}$, to yield the new memory $\bm{H}_t$. In \S\ref{subsec:ablation}, we show that the vanilla ConvGRU has outperformed the simple point cloud merging~\cite{caesar2019nuScenes} and the temporal 3D ConvNet~\cite{luo2018fast}. Next, we present how we promote the vanilla ConvGRU with the STA and TTA modules.

\noindent\textbf{Spatial Transformer Attention.} The core idea of the STA module is to attend each pixel-level feature $\bm{x}\in\bm{X}_t$ with a rich spatial context, to better distinguish a foreground object from the background noise. Basically, a transformer attention receives a query $\bm{x}_q\in\bm{X}_t$ and a set of keys $\bm{x}_k\in\bm{\Omega}_{\bm{x}_q}$ (\eg, the neighbors of $\bm{x}_q$), to calculate an attentive output $\bm{y}_q$. The STA is designed as an intra-attention, which means both the query and key are from the same input feature $\bm{X}_t$. 

Formally, given a query $\bm{x}_q\in\bm{X}_t$ at location $q\in{w\times{h}}$, the attentive output $\bm{y}_q$ is computed by:
\begin{equation}
\begin{aligned}
\label{eq:transformer}
\bm{y}_q = \sum_{k\in\bm{\bm{\Omega}}_q}{A(\phi_Q(\bm{x}_q), \phi_K(\bm{x}_k))\circ\phi_V(\bm{x}_k)},
\end{aligned}
\end{equation}
where $A(\cdot, \cdot)$ is the attention weight. $\phi_K$, $\phi_Q$ and $\phi_V$ are the linear layers that map the inputs $\bm{x}_q, \bm{x}_k\in\bm{X}_t$ into different embedding subspaces. The attention weight $A(\cdot, \cdot)$ is computed from the embedded query-key pair $(\phi_Q(\bm{x}_q), \phi_K(\bm{x}_k))$, and is then applied to the neighbor values $\phi_V(\bm{x}_k)$. 

Since we need to obtain the attention for all the query-key pairs, the linear layers, $\phi_K$, $\phi_Q$ and $\phi_V$, are then achieved by the convolutional layers, $\Phi_K$, $\Phi_Q$ and $\Phi_V$, to facilitate the computation. Specifically, the input features $\bm{X}_t$ are first embedded as $\bm{K}_t$, $\bm{Q}_t$ and $\bm{V}_t\in\mathbb{R}^{w\times h\times c'}$ through $\Phi_K$, $\Phi_Q$ and $\Phi_V$. Then, we adjust the tensor shapes of $\bm{K}_t$ and $\bm{Q}_t$ to $l \times c'$, where $l=w\times h$, in order to compute the attention weight:
\begin{equation}
\begin{aligned}
\tilde{\bm{A}} = \text{softmax}(\bm{Q}_t\cdot\bm{K}_t^T)\in\mathbb{R}^{l\times l}
\end{aligned}
\end{equation}
where $A(\cdot, \cdot)$ is realized as a softmax layer to normalize the attention weight matrix. After that, $\tilde{\bm{A}}$ is employed to aggregate information from the values $\bm{V}_t$ through a matrix multiplication, generating the attentive output $\tilde{\bm{A}}\cdot\bm{V}_t$, with the tensor shape recovered to $w\times h \times c'$. Finally, we obtain the spatially enhanced features $\bm{X}_t^{'}$ through a residual operation~\cite{he2016deep}, which can be summarized as:
\begin{equation}
\begin{aligned}
\label{eq:spatial}
\bm{X}_t^{'} = \bm{W}_\text{out}*{(\tilde{\bm{A}}\cdot\bm{V}_t)} + \bm{X}_t\in\mathbb{R}^{w\times h\times c}
\end{aligned}
\end{equation}
where $\bm{W}_\text{out}$ is the output layer of the attention head that maps the embedding subspace ($c'$-dim) of $\tilde{\bm{A}}\cdot\bm{V}_t$ back to the original space ($c$-dim). In this way, $\bm{X}_t^{'}$ contains the information from its spatial context and thus can better focus on the meaningful foreground objects.

\noindent\textbf{Temporal Transformer Attention.} 
To adaptively align the features of dynamic objects from $\bm{H}_{t-1}$ to $\bm{X}_t^{'}$, we apply the modified deformable convolutional layers~\cite{zhu2019deformable,zhu2019empirical} as a special instantiation of the transformer attention. The core is to attend the queries in $\bm{H}_{t-1}$ with adaptive supporting key regions computed by integrating the motion information. 

Specifically, given a vanilla deformable convolutional layer with kernel size $3\times 3$, let $\bm{w}_m$ denotes the learnable weights, and $p_m\in \{(-1,-1), (-1,0), ..., (1,1)\}$ indicates the predetermined offset in total $M=9$ grids. The output $\bm{h}_q^{'}$ for input $\bm{h}_{q}\in{\bm{H}_{t-1}}$ at location $q\in{w\times{h}}$ can be expressed as:
\begin{equation}
\begin{aligned}
\label{dcn}
\bm{h}_q^{'} = \sum_{m=1}^{M}\bm{w}_m\cdot{\bm{h}_{q+p_m+\Delta{p_m}}},
\end{aligned}
\end{equation}
where $\Delta{p_m}$ is the deformation offset learnt through a separate regular convolutional layer $\Phi_R$, \ie, $\Delta{p_m}\in\Delta{\bm{P}_{t-1}}=\Phi_R(\bm{H}_{t-1})\in\mathbb{R}^{w\times{h}\times{2r^{2}}}$, where the channel number $2r^{2}$ denotes the offsets in the x-y plane for the $r\times{r}$ convolutional kernel. We can also reformulate Eq.~\ref{dcn} from the perspective of transformer attention as in Eq.~\ref{eq:transformer}, such that the attentive output $\bm{h}_q^{'}$ of query $\bm{h}_q$ is given by:
\begin{equation}
\begin{aligned}
\label{eq:tta}
\bm{h}_q^{'} = \sum_{m=1}^{M}\bm{w}_m\cdot\sum_{k\in\bm{\Omega}_q}G(k, q+p_m+\Delta p_m)\cdot \phi_\text{V}(\bm{h}_k),
\end{aligned}
\end{equation}
where $\phi_\text{V}$ is an identity function, and $\bm{w}_m$ acts as the weights in different attention heads~\cite{zhu2019empirical}, with each head corresponding to a sampled key position $k\in{\bm{\Omega}_q}$. $G(\cdot, \cdot)$ is the attention weight defined by a bilinear interpolation function, such that $G(a, b) = max(0, 1-|a-b|)$.

The supporting key regions ${\bm{\Omega}_q}$ play an important role in attending $\bm{h}_q$, which are determined by the deformation offset $\Delta{p_m}\in\Delta{\bm{P}_{t-1}}$. In our TTA module, we compute $\Delta{\bm{P}_{t-1}}$ not only through $\bm{H}_{t-1}$, but also through a \textit{motion map}, which is defined as the difference of $\bm{H}_{t-1}$ and $\bm{X}_t^{'}$:
\begin{equation}
\begin{aligned}
\label{eq:offset}
{\Delta\bm{P}_{t-1}} = \Phi_{R}([\bm{H}_{t-1}, \bm{H}_{t-1}- \bm{X}_t^{'}])\in\mathbb{R}^{w\times{h}\times{2r^{2}}},
\end{aligned}
\end{equation}
where $\Phi_{R}$ is a regular convolutional layer with the same kernel size as that in the deformable convolutional layer, and $[\cdot, \cdot]$ is the concatenation operation. The intuition is that, in the \textit{motion map}, the features response of the static objects is very low since they have been spatially aligned in $\bm{H}_{t-1}$ and $\bm{X}_t^{'}$, while the features response of the dynamic objects remains high. Therefore, we integrate $\bm{H}_{t-1}$ with the \textit{motion map}, to further capture the motions of dynamic objects. 
Then, $\Delta{\bm{P}_{t-1}}$ is used to select the supporting key regions and further attend $\bm{H}_{t-1}$ for all the query regions $q\in{w\times{h}}$ in terms of Eq.~\ref{eq:tta}, yielding a temporally attentive memory $\bm{H}_{t-1}^{'}$. Since the supporting key regions are computed from both $\bm{H}_{t-1}$ and $\bm{X}_t^{'}$, our TTA module can be deemed as an inter-attention.

Additionally, we can stack multiple modified deformable convolutional layers to get a more accurate $\bm{H}_{t-1}^{'}$. In our implementation, we adopt two layers. The latter layer takes as input $[\bm{H}_{t-1}^{'}, \bm{H}_{t-1}^{'}- \bm{X}_t^{'}]$ to predict the deformation offset according to Eq.~\ref{eq:offset}, and the offset is then used to attend $\bm{H}_{t-1}^{'}$ via Eq.~\ref{eq:tta}. Accordingly, we can now utilize the temporally attentive memory $\bm{H}_{t-1}^{'}$ and the spatially attentive input $\bm{X}_t^{'}$ to compute the new memory $\bm{H}_{t}$ in the recurrent unit (see Fig.~\ref{fig:GRU}). Finally, a RPN detection head is applied on $\bm{H}_{t}$ to produce the final detection results $\bm{Y}_{t}$. 

\vspace{-2pt}
\subsection{Network Details}
\vspace{-1pt}
\label{subsec:detail}
\noindent\textbf{PMPNet.} Our PMPNet is an end-to-end differentiable model achieved by parameterizing all the functions with neural networks. Given a discretized point cloud frame $\bm{I}_{t}\in{\mathbb{R}^{P\times{N} \times{D}}}$ with $P$ pillar nodes, $F_\text{PFN}$ is first used to generates the initial node state $G^{0}\in{\mathbb{R}^{P\times{L}}}$ for all the nodes (Eq.~\ref{eq:pfn}), which is realized by a $1\times{1}$ convolutional layer followed by a max pooling layer that operates on the $N$ points. In each iteration step $s$, the edge features from the $K$ neighbor nodes are first collected as $\tilde{G}^{s}\in{\mathbb{R}^{P\times{K}\times{2L}}}$ (Eq.~\ref{eq:edge}) with a concatenation operation. Then the message functions (Eq.~\ref{eq:message1} and Eq.~\ref{eq:message2}) map the collected features $\tilde{G}^{s}$ to $M^{s}\in{\mathbb{R}^{P\times{L^{'}}}}$, through a $1\times{1}$ convolutional layer followed by a max pooling layer performing on the $K$ messages. The update function (Eq.~\ref{eq:update}) then updates the node state using a GRU with fully connected layers, by considering both the $G^{s}\in{\mathbb{R}^{P\times{L}}}$ and $M^{s}\in{\mathbb{R}^{P\times{L^{'}}}}$, and outputs $G^{s+1}\in{\mathbb{R}^{P\times{L}}}$. After $S$ iteration steps, we get the final node state $G^{S}\in{\mathbb{R}^{P\times{L}}}$, and scatter it back to a 3D tensor $\bm{\tilde{I}}_{t\in{\mathbb{R}^{W\times{H\times{L}}}}}$ (Eq.~\ref{eq:2dbackbone}).
\begin{table*}
\centering
\setlength\tabcolsep{4pt}
\renewcommand\arraystretch{1.00}
\resizebox{0.90\textwidth}{!}{
\begin{tabular}{l||ccccccccccc}
\toprule[1pt]
\vspace{-1.5pt}
  ~~~Method   & \multicolumn{1}{c}{Car}   & \multicolumn{1}{c}{Pedestrian}   & \multicolumn{1}{c}{Bus}        & \multicolumn{1}{c}{Barrier}        & \multicolumn{1}{c}{T.C.}      & \multicolumn{1}{c}{Truck}       & \multicolumn{1}{c}{Trailer}   & \multicolumn{1}{c}{Moto. }  & \multicolumn{1}{c}{Cons.}   & \multicolumn{1}{c}{Bicycle}  & Mean  \\ \hline
            
 VIPL\_ICT~\cite{leaderboard}	& 71.9  &57.0 & 34.1   & 38.0   & 27.3     & 20.6  & 26.9 &20.4 &3.3 &0.0 & 29.9   \\
            
MAIR~\cite{simonelli2019disentangling}	& 47.8  &37.0 & 18.8   & 51.1   & 48.7     & 22.0  & 17.6 &29.0 &7.4 &\textbf{24.5} & 30.4   \\
            
PointPillars~\cite{lang2019pointpillars}	& 68.4  &59.7 & 28.2   & 38.9   & 30.8     & 23.0  & 23.4 &27.4 &4.1 &1.1 & 30.5   \\   
 
SARPNET~\cite{ye2020sarpnet}  & 59.9  &69.4 & 19.4   & 38.3   & 44.6  &18.7 &18.0 &29.8 & 11.6 & {14.2}  & 32.4   \\ 
 
WYSIWYG~\cite{hu2019you}	& 79.1  &65.0 & 46.6   & 34.7   & 28.8     & 30.4  & 40.1 &18.2 &7.1 &0.1 & 35.0   \\ 
 
Tolist~\cite{leaderboard}	&79.4  &71.2 & 42.0   & \textbf{51.2}   & 47.8    &\textbf{34.5} &34.8 &36.8 &9.8    & 12.3  & 42.0    \\ \hline
 
 
 \textbf{Ours}  & \textbf{79.7}  &\textbf{76.5} & \textbf{47.1}   & 48.8  & \textbf{58.8}                        & 33.6  & \textbf{43.0}  & \textbf{40.7}& \textbf{18.1} & 7.9 & \textbf{45.4}  \\ 
\bottomrule[1pt]
\end{tabular}%
}
\caption{\small {\textbf{Quantitative detection results on the nuScenes 3D object detection benchmark.} T.C. presents the traffic cone. Moto. and Cons. are short for the motorcycle and construction vehicle, respectively. Our 3D \textit{video} object detector outperforms the \textit{single-frame} detectors, achieving state-of-the-art performance on the leaderboard. }}
\label{tb:mAP}
\vspace{-4mm}
\end{table*}

\noindent\textbf{Backbone Module.}
As in~\cite{zhou2018voxelnet}, we utilize a 2D backbone network to further extract features for $\bm{\tilde{I}}_{t\in{\mathbb{R}^{W\times{H\times{L}}}}}$, which consists of three blocks of fully convolutional layers. Each block is defined as a tuple $(S, Z, C)$. All the blocks have $Z\times{Z}$ convolutional kernels with output channel number $C$. The first layer of each block operates at stride $S$, while other layers have stride 1. The output features of each block are resized to the same resolution via upsampling layers and then concatenated together to merge the semantic information from different feature levels. 

\noindent\textbf{AST-GRU Module.} In our STA module, all the linear functions in Eq.~\ref{eq:transformer} and $\bm{W}_\text{out}$ in Eq.~\ref{eq:spatial} are $1\times{1}$ convolution layers. In our TTA module, the regular convolutional layers, the deformable convolutional layers and the ConvGRU all have learable kernels of size $3\times{3}$.

\noindent\textbf{Detection Head.} The detection head in~\cite{zhou2018voxelnet} is applied on the attentive memory features. In particular, the smooth L1 loss and the focal loss~\cite{lin2017focal} count for the object bounding box regression and classification, respectively. A corss-entropy loss is used for the orientation classification. For the velocity regression required by the nuScenes benchmark, a simple L1 loss is adopted and shows substantial results.

\vspace{-1.5mm}
\section{Experimental Results}
\vspace{-1pt}

\noindent\textbf{3D Video Object Detection Benchmark.} We evaluate our algorithm on the challenging nuScenes 3D object detection benchmark~\cite{caesar2019nuScenes}, since the KITTI benchmark~\cite{geiger2012we} does not provide the point cloud videos. nuScenes is a large-scale dataset with a total of 1,000 scenes, where 700 scenes (28,130 samples) are for training and 150 scenes (6,008 samples) are for testing, resulting 7$\times$ as many annotations as the KITTI. The samples (also named as keyframes) in each video are annotated every \SI{0.5}{s} with a full 360-degree view, and their point clouds are densified by the 10 non-keyframe sweeps frames, yielding around {300,000} point clouds with 5-dim representation $(x, y, z, r, \Delta t)$, where $r$ is the reflectance and $\Delta t$ describes the time lag to the keyframe (ranging from \SI{0}{s} to \SI{0.45}{s}). Besides, nuScenes requires detecting objects for 10 classes with full 3D boxes, attributes and velocities.

\noindent\textbf{Implementation Details.} For each keyframe, we consider the point clouds within range of $[-50, 50] \times [-50, 50]\times [-5, 3]$ meters along the X, Y and Z axes. The pillar resolution on the X-Y plane is $0.25^2~m^2$. The pillar number $P$ used in PMPNet is 16,384, sampled from the total 25,000 pillars, with each pillar containing most $N=60$ points. The input point cloud is a $D=5$ dimensions representation $(x,y,z,r,\Delta t)$, which are then embedded into $L=L^{'}=64$ dimensions feature space after the total $S=3$ graph iteration steps. The convolutional kernels in the 2D backbone are of size  $Z=3$ and the output channel number $C$ in each block is $(64, 128, 256)$. The upsampling layer has kernel size 3 and channel number 128. Thus, the final features map produced by the 2D backbone has a resolution of $100\times 100\times 384$. We calculate anchors for different classes using the mean sizes and set the matching threshold according to the class instance number. 
The coefficients of the loss functions for classification, localization and velocity prediction are set to 1, 2 and 0.1, respectively. NMS with IOU threshold 0.5 is utilized when generating the final detections. 
In both training and testing phases, we feed most 3 consecutive keyframes to the model due to the memory limitation. The training procedure has two stages. In the first stage, we pre-train the spatial features encoding component using the one-cycle policy~\cite{smith2017cyclical} with a maximum learning rate of 0.003. 
Then, we fix the learning rate to 0.0002 in the second stage to train the full model. We train 50 epochs for both stages with batch size 3. Adam optimizer~\cite{kingma2015adam} is used to optimize the loss functions. 

\vspace{-2mm}
\subsection{Quantitative and Qualitative Performance}
\begin{figure}
\vspace{-3mm}
\centering     
\subfigure[Detection results from the single-frame 3D object detector~\cite{lang2019pointpillars}.]{\includegraphics[width=0.496\textwidth]{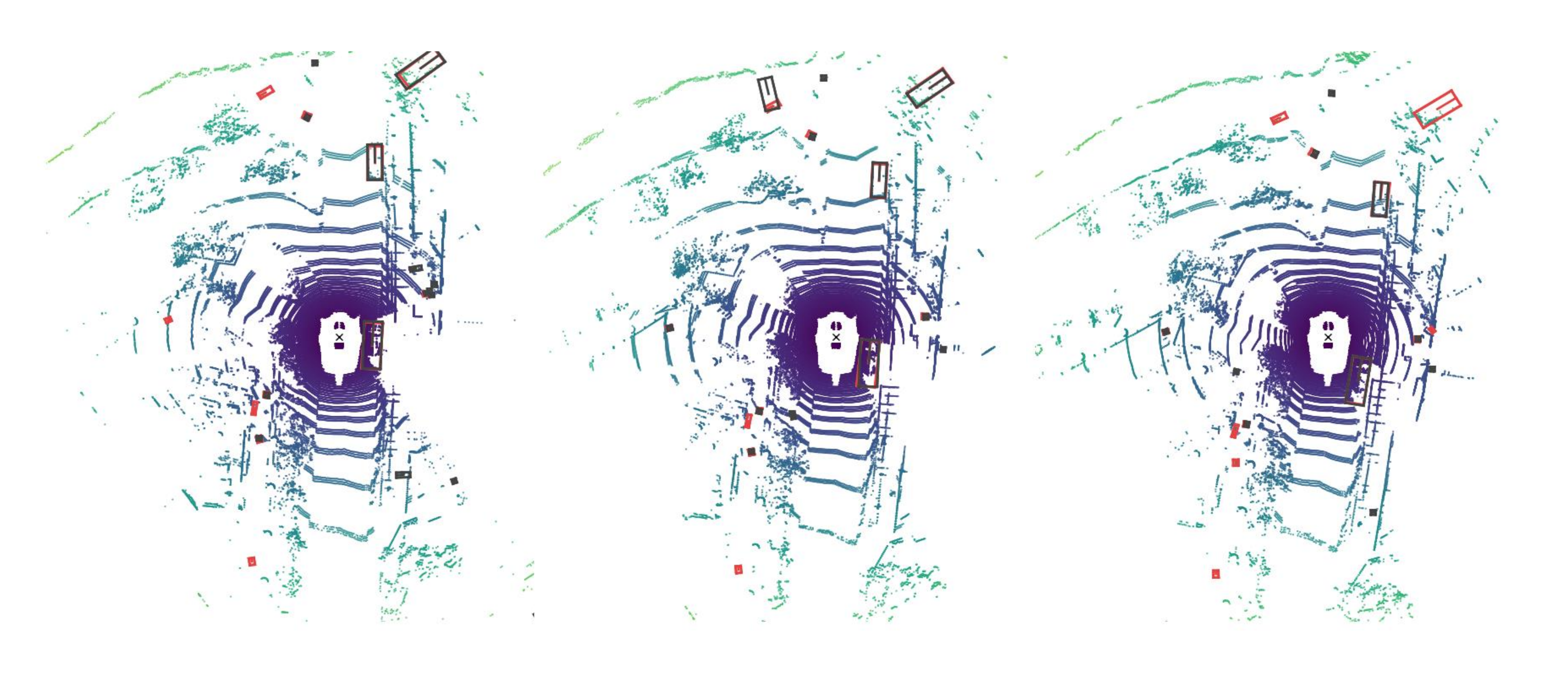}}
\subfigure[Detection results from our 3D video object detector.]{\includegraphics[width=0.496\textwidth]{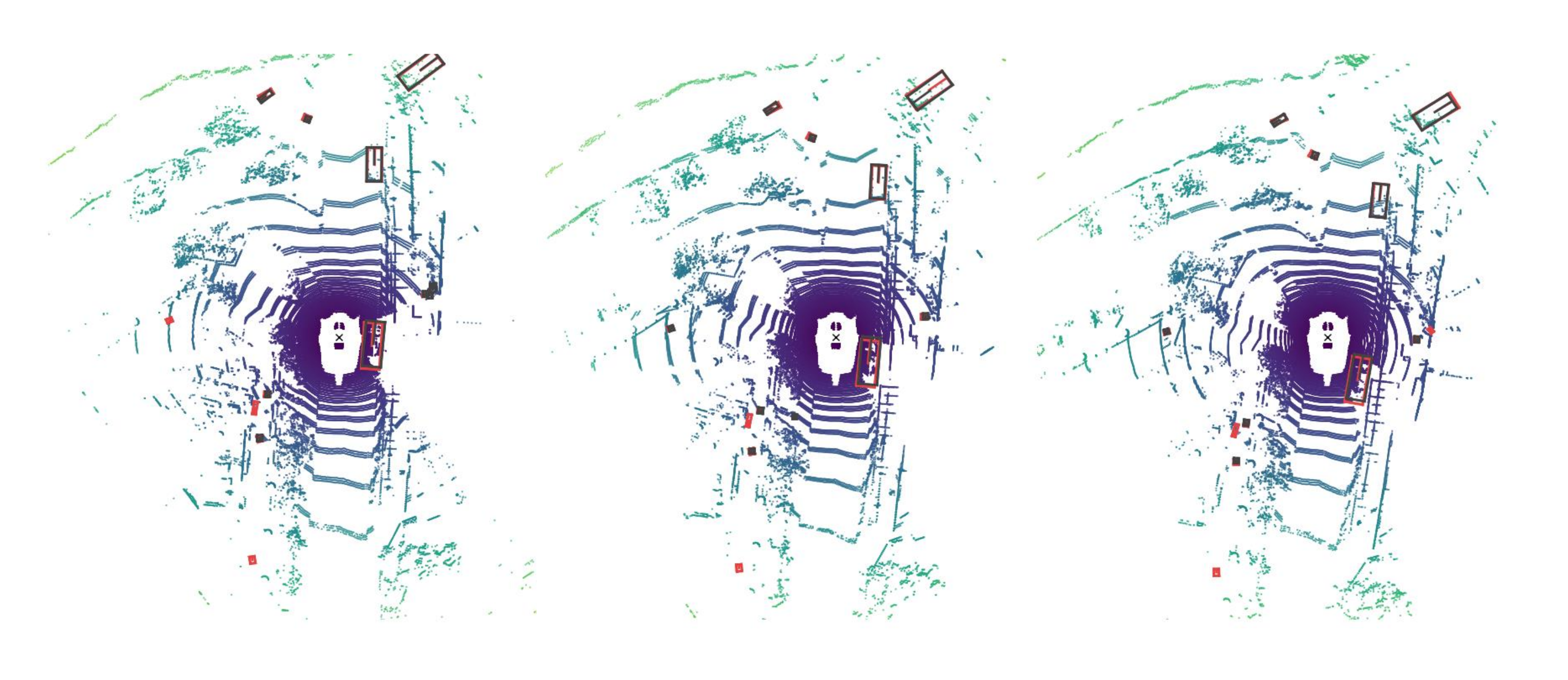}}

\caption{\small {\textbf{Detections for the distant cars.} The grey and red boxes indicate the predictions and ground-truths, respectively.}}
\label{fig:distant}
\vspace{-4mm}
\end{figure}

\vspace{-1mm}
\label{subsec:quantitative}
We present the performance comparison of our algorithm and other state-of-the-art approaches on the nuScenes benchmark in Table~\ref{tb:mAP}. PointPillars~\cite{lang2019pointpillars}, SARPNET~\cite{ye2020sarpnet}, WYSIWYG~\cite{hu2019you} and Tolist~\cite{leaderboard} are all voxel-based single-frame 3D object detectors. In particular, PointPillars is used as the baseline of our model. WYSIWYG is a recent algorithm that extends the PointPillars with a voxelized visibility map. Tolist uses a multi-head network that contains multiple prediction heads for different classes.
Our 3D video object detector outperforms these approaches by a large margin. In particular, we improve the official PointPillars algorithm by 15\%. Please note that there is a severe class imbalance issue in the nuScenes dataset. The approach in~\cite{zhu2019class} designs a class data augmentation algorithm. Further integrating with these techniques can promote the performance of our model. But we focus on exploring the spatiotemporal coherence in the point cloud video, and handling the class imbalance issue is not the purpose in this work. In addition, we further show some qualitative results in Fig.~\ref{fig:distant}. Besides the occlusion situation in Fig.~\ref{fig:video}, we present another case of detecting the distant car (the car on the top right), whose point clouds are especially sparse, which is very challenging for the single-frame detectors. Again, our 3D video object detector effectively detects the distant car using the attentive temporal information.

\vspace{-1mm}
\subsection{Ablation Study}
\vspace{-1mm}
\label{subsec:ablation}
In this section, we investigate the effectiveness of each module in our algorithm. Since the training samples in the nuScenes is 7$\times$ as many as those in the KITTI (28,130 vs 3,712), it is non-trivial to train multiple models on the whole dataset. Hence, we use a mini train set for validation purposes. It contains around 3,500 samples uniformly sampled from the original train set. Besides, PointPillars~\cite{lang2019pointpillars} is used as the baseline detector in our model. 

First, we evaluate our PMPNet in the spatial feature encoding component, by replacing the PFN in PointPillars with PMPNet. As shown in Table~\ref{tb:ablation}, it improves the baseline by 2.05\%. Second, we validate the ability of each module in the spatiotemporal feature aggregation component (\ie, ConvGRU, STA-GRU and TTA-GRU) through adding each module to the PointPillars. We can see that all these modules achieve better performance than the single-frame detector. Moreover, we compare AST-GRU with other video object detectors. Since each keyframe in nuScenes contains point clouds merged by previous 10 non-keyframe sweeps, the PointPillars baseline, trained on the merged keyframes, can be deemed as the simplest video object detector. Our AST-GRU improves it by 5.98\%. Then, we compare AST-GRU with the temporal 3D ConvNet-based method by implementing the late feature fusion module in~\cite{luo2018fast}. Temporal 3D ConvNet can only access a single keyframe label (\SI{0.5}{s}) during training, and aggregating more labels instead impairs the performance. According to Table~\ref{tb:ablation}, the 3D ConvNet-based method surpasses the PointPillars baseline by 1.82\%. In contrast, our AST-GRU further outperforms it by 4.16\%, which demonstrates the importance of the long-term temporal information. Finally, the full model with the PMPNet achieves the best performance.

Finally, we analyze the effect of the input sequence length. Since each keyframe contains quantities of point clouds that will increase the memory demand, we conduct this experiment without using the point clouds in non-keyframe sweeps. Experimental results with different input lengths are shown in Table.~\ref{tb:sequence}, which demonstrates that using the previous long-term temporal information (\SI{2.5}{s}) can consistently gain better performance in 3D object detection.
\begin{table}
\centering
\renewcommand\arraystretch{1.00}
\resizebox{0.465\textwidth}{!}{
\begin{tabular}{c|c|l|cc}
\hline
\multirow{2}{*}{Components}                                                         & \multicolumn{2}{c|}{\multirow{2}{*}{Modules}} & \multicolumn{2}{c}{Performance} \\
                                                                                    & \multicolumn{2}{c|}{}                         & mAP             & $\Delta$       \\ \hline
\multirow{2}{*}{\begin{tabular}[c]{@{}c@{}}Single-frame\\ 3D Object Detector\end{tabular}} & \multicolumn{2}{c|}{PointPillars (PP) }         & 21.30           & -              \\
                                                                                    & \multicolumn{2}{c|}{PP + PMPNet}              & 23.35           & +2.05          \\ \hline
\multirow{6}{*}{\begin{tabular}[c]{@{}c@{}}3D Video\\ Object Detector\end{tabular}}        & \multicolumn{2}{c|}{PP + 3D ConvNet}            & 23.12           & +1.82          \\
                                                                                    & \multicolumn{2}{c|}{PP + ConvGRU}               & 23.83           & +2.53          \\
                                                                                    & \multicolumn{2}{c|}{PP + STA-GRU}               & 25.23           & +3.93          \\
                                                                                    & \multicolumn{2}{c|}{PP + TTA-GRU}               & 25.32           & +4.02          \\
                                                                                    & \multicolumn{2}{c|}{PP + AST-GRU}               & 27.28           & +5.98          \\ \cline{2-5} 
                                                                                    & \multicolumn{2}{c|}{\textbf{Full Model}}      & \textbf{29.35}  & \textbf{+8.05} \\ \hline
\end{tabular}
}
\caption{\small {\textbf{Ablation study for our 3D video object detector.} PointPillars~\cite{lang2019pointpillars} is the reference baseline for computing the relative improvement ($\Delta$).}
}
\label{tb:ablation}
\vspace{-2mm}
\end{table}

\begin{table}
\centering
\renewcommand\arraystretch{1.00}
\resizebox{0.38\textwidth}{!}{
\begin{tabular}{c|l|c|l|cc}
\hline
\multicolumn{2}{c|}{\multirow{2}{*}{Aspect}}                                                                   & \multicolumn{2}{c|}{\multirow{2}{*}{Modules}} & \multicolumn{2}{c}{Performance}      \\
\multicolumn{2}{c|}{}                                                                                          & \multicolumn{2}{c|}{}                         & mAP                        & $\Delta$ \\ \hline
\multicolumn{2}{c|}{\multirow{5}{*}{\begin{tabular}[c]{@{}c@{}}Input Lengths\\ (Full Model)\end{tabular}}} & \multicolumn{2}{c|}{T=1}                      & \multicolumn{1}{c|}{16.84} & -    \\ \cline{3-6} 
\multicolumn{2}{c|}{}                                                                                          & \multicolumn{2}{c|}{T=2}                      & \multicolumn{1}{c|}{19.34} & +2.50    \\ \cline{3-6} 
\multicolumn{2}{c|}{}                                                                                          & \multicolumn{2}{c|}{{T=3}}                      & \multicolumn{1}{c|}{{20.27}} & +3.43   \\ \cline{3-6} \multicolumn{2}{c|}{}                                                                                          & \multicolumn{2}{c|}{{T=4}}                      & \multicolumn{1}{c|}{{20.77}} & +3.93  \\ \cline{3-6} \multicolumn{2}{c|}{}                                                                                          & \multicolumn{2}{c|}{\textbf{T=5}}                      & \multicolumn{1}{c|}{\textbf{21.52}} & \textbf{+4.68}    \\
 \hline
\end{tabular}
}
\caption{\small {\textbf{Ablation study for the input lengths.} Detection results with one input frame are used as the reference baseline.
}}
\label{tb:sequence}
\vspace{-3mm}
\end{table}

\vspace{-4mm}
\section{Conclusion}
\vspace{-1mm}
This paper proposed a new 3D video object detector for exploring the spatiotemporal information in point cloud video. It has developed two new components: spatial feature encoding component and spatiotemporal feature aggregation component. We first introduce a novel PMPNet that considers the spatial features of each point cloud frame. PMPNet can effectively enlarge the receptive field of each pillar grid through iteratively aggregating messages on a $k$-NN graph. Then, an AST-GRU module composed of STA and TTA is presented to mine the spatiotemporal coherence in consecutive frames by using an attentive memory gating mechanism. The STA focuses on detecting the foreground objects, while the TTA aims to align the dynamic objects. Extensive experiments on the nuScenes benchmark have proved the better performance of our model.

\newpage

{\small
\bibliographystyle{ieee_fullname}
\bibliography{detection}
}

\end{document}